\documentclass[twoside]{article}

\usepackage[round]{natbib}
\usepackage{mathrsfs}
\usepackage[accepted]{aistats2025}
\usepackage{url}

\usepackage{amsfonts}
\usepackage{booktabs}
\usepackage{amsthm}

\usepackage{algorithm}
\usepackage{algpseudocode}
\usepackage{graphicx}
\usepackage{subcaption}
\usepackage{soul}
\DeclareMathOperator*{\argmin}{argmin}

\newtheorem{theorem}{Theorem}[section]
\newtheorem{lemma}{Lemma}[section]
\newtheorem{corollary}{Corollary}[section]

\begin{document}

%

%

\twocolumn[

\aistatstitle{Post-processing for Fair Regression via Explainable SVD}

\aistatsauthor{ Zhiqun Zuo \And Ding Zhu \And Mohammad Mahdi Khalili }

\aistatsaddress{ The Ohio State University \And  The Ohio State University \And The Ohio State University } ]

\begin{abstract}
    This paper presents a post-processing algorithm for training fair neural network regression models that satisfy statistical parity, utilizing an explainable singular value decomposition (SVD) of the weight matrix. We propose a linear transformation of the weight matrix, whereby the singular values derived from the SVD of the transformed matrix directly correspond to the differences in the first and second moments of the output distributions across two groups. Consequently, we can convert the fairness constraints into constraints on the singular values. We analytically solve the problem of finding the optimal weights under these constraints. Experimental validation on various datasets demonstrates that our method achieves a similar or superior fairness-accuracy trade-off compared to the baselines without using the sensitive attribute at the inference time.\footnote{The code for this paper can be found in \url{https://github.com/osu-srml/svd_fairness.git}}
\end{abstract}

\section{INTRODUCTION}
Machine learning models are increasingly central to human-centric applications, raising significant concerns about their impact on social fairness \citep{liu2024evaluating}. These models have been shown to manifest biases against specific groups in several notable instances. For example, the COMPAS recidivism prediction tool was found to be biased against African Americans \citep{dieterich2016compas}. In another case, an Amazon recruitment tool that assessed applicants based on their resumes was shown to yield less favorable outcomes for women, a discrepancy attributed to their underrepresentation in technical roles \citep{wicksautomated}. Additionally, a study within a US healthcare system indicated that black patients were only assigned the same risk level as white patients when they exhibited more severe symptoms \citep{obermeyer2019dissecting}. Even sophisticated large language models, such as ChatGPT, have been found to perpetuate gender stereotypes \citep{gross2023chatgpt}.

To effectively address unfairness in machine learning models, it is crucial to first establish a definition of fairness. There are various definitions for fairness in the literature and they typically fall into two categories: group fairness and individual fairness \citep{verma2018fairness}. Group fairness encompasses concepts such as equalized odds \citep{romano2020achieving}, equal opportunity \citep{shen2022optimising}, and statistical parity \citep{jiang2022generalized}. On the other hand, individual fairness includes definitions such as fairness through awareness \citep{dwork2012fairness} and counterfactual fairness \citep{kusner2017counterfactual}.

There are various techniques to satisfy a fairness notion which can be broadly categorized into three types: pre-processing, in-processing, and post-processing methods. Pre-processing methods \citep{d2017conscientious,pmlr-v238-abroshan24a,abroshan2022counterfactual,NEURIPS2023_2828ee0c} aim to modify the training data to facilitate the learning of a fair model. In-processing methods \citep{wan2023processing,pmlr-v202-khalili23a} alter the training procedures (e.g., the objective function) to ensure fairness. Post-processing methods \citep{kim2019multiaccuracy,NEURIPS2021_ed277964}, on the other hand, directly modify the model's predictions to achieve fairness. 

Among these methods, post-processing has the advantage of speed as it does not require retraining the model. This is particularly significant given that training large-scale models is becoming increasingly expensive today. 
Unlike fair classification, fair regression problems have received less attention in the literature. \cite{agarwal2019fair} propose an in-processing algorithm called reduction approach to fair regression by reducing the constrained optimization problem to a standard and unconstrained regression problem. In addition, post-processing methods have been adapted for fair regression problems \citep{xian2024differentially, chzhen2020fair}. These post-processing methods primarily focus on mapping an unfair output distribution to a fair one without modifying the learned model, which limits their effectiveness. Additionally, these methods require access to sensitive attributes at the time of inference, which may not be feasible in some real-world scenarios. Furthermore, the post-processing methods for regression are required to discretize the target distribution, implying that the target values must be bounded. The choice of discretization width also poses a significant challenge as it can greatly impact the performance of the model.

In this work, we introduce a method for post-processing the model weights of a pre-trained regression model by eliminating unfair factors presented in the weights of neural network models. Drawing on insights from the recent study by \cite{wang2024svd}, our aim is to establish a direct link between singular values and singular vectors of some transformation of the weight matrices and unfairness. 

Leveraging this connection, we introduce a novel approach to obtain a revised weight matrix by modifying the singular values to satisfy specific constraints. Our goal is to preserve the output distribution of the original model while making these adjustments. We demonstrate that with these constraints, the problem of improving fairness can be transformed into a convex optimization problem, which admits a closed-form solution. Subsequently, we provide an Explainable SVD based Fairness enhancement (ESVDFair) algorithm to post-process pre-trained neural networks. The processed model can then perform fair inference without requiring any sensitive attributes during the inference stage.

The subsequent sections of this paper are organized as follows. Section 2 introduces the preliminaries essential for understanding the methods discussed, including the definition of statistical parity in the context of regression. Section 3 elucidates the connection between the explainable SVD and statistical parity, providing a theoretical foundation for the techniques developed in this work. Section 4 details how we transform the problem of fairness enhancement through post-processing into a convex optimization problem. Section 5 validates our algorithm through empirical experiments conducted on two distinct datasets, demonstrating its effectiveness in real-world scenarios. Finally, Section 6 concludes the paper with a summary of our findings.


\section{PRELIMINARIES}
Consider a regression model $f:\mathcal{X}\to \mathcal{Y}$ where $\mathcal{X}$ is the feature space, and $\mathcal{Y}$ is the output space. Each feature vector $\mathscr{X} \in \mathcal{X}$ is associated with an output $\mathscr{Y}$ and a binary sensitive attribute $\mathscr{A}\in \{1, 2\}$.\footnote{$\mathscr{X},\mathscr{A},\mathscr{Y}$ are random variables. Their realizations are denoted by $x, a, y$, respectively.} We denote the output of $f$ by $\hat{\mathscr{Y}} = f(\mathscr{X})$. In our setting, $\mathscr{X}$ may or may not include the sensitive attribute $\mathscr{A}$. If $\mathscr{X}$ does not include $\mathscr{A}$, the predictor is called attribute unaware.  Predictor $f$ can exhibit biases and unfairness against a sensitive group, and our goal is to mitigate such biases. To measure fairness, we employ the notion of statistical parity \citep{dwork2012fairness} and try to modify predictor $f$ to satisfy this notion. 
Statistical parity requires the output distribution to be independent of the sensitive attribute $\mathscr{A}$. In other words, $\mathscr{\hat{Y}}$ satisfies statistical parity, if the following condition holds:
\begin{equation}\label{eq:sp-definition}
    \Pr \{\mathscr{\hat{Y}} = y | \mathscr{A} = a\} = \Pr \{\mathscr{\hat{Y}} = y\}, ~~ \forall y \in \mathcal{Y}, \forall a \in \{1, 2\}.
\end{equation}
When the domain $\mathcal{Y}$ is continuous, it is hard to quantitatively measure the violation of statistical parity. 
In this paper, we utilize the $\alpha$-approximate statistical parity \citep{xian2024differentially} to measure the fairness violation. A regressor satisfies $\alpha$-approximate statistical parity if the following condition holds:
\begin{equation}\label{eq:ks}
\resizebox{\linewidth}{!}{$
    \sup_{t \in \mathbb{R}}\left|\int_{-\infty}^{t} \left(\Pr(\mathscr{\hat{Y}} = y|\mathscr{A} = a_{1}) - \Pr(\mathscr{\hat{Y}} = y|\mathscr{A} = a_{2})\right)\mathrm{d}y\right| \leq \alpha.
$}
\end{equation}
The definition utilize the Kolmogorov-Smirnov (KS) distance of two distributions. In real experiments, the probability density function could be estimated by discretizing $\mathcal{Y}$ into bins.\footnote{It should be noticed that the discretization is only used for evaluation. Unlike previous post-processing methods, our proposed algorithm involves no discretization.} The smaller $\alpha$ value implies a better fairness level. 

In this paper, we consider a \textit{pre-trained} model $f$ in the form of a neural network with $L$ layers. We assume that the layer $l$ has the weight matrix $W^{[l]} \in \mathbb{R}^{m^{[l]} \times n^{[l]}}$. We denote the input of layer $l$ by $\mathscr{X}^{[l]}$. If the input is associated with group $a$, then we use notation $\mathscr{X}^{[l]}_a$. Note that, $\mathscr{X}^{[l+1]} = \sigma(\mathscr{X}^{[l]} (W^{[l]})^T), l<L$, and $\hat{\mathscr{Y}} = (\mathscr{X}^{[L]} (W^{[L]})^T)$, where $\sigma(.)$ is a non-linear function. We let $\mathscr{Z}_a^{[l]} =  \mathscr{X}_{a} (W^{[l]})^T$, and $\mathscr{Z}^{[l]} =  \mathscr{X}^{[l]} (W^{[l]})^T$.

Our goal is to modify the matrix $W^{[l]}$ to ensure statistical parity with minimal performance reduction. More precisely, we will propose an algorithm to find matrix $W^{'[l]}$ similar to $W^{[l]}$  such that random variables $\mathscr{Z}_{1}'^{[l]} = \mathscr{X}_{1}^{[l]} (W^{'[l]})^T$ and $\mathscr{Z}_{2}'^{[l]} = \mathscr{X}_{2}^{[l]} (W^{'[l]})^T$ have the same first and second moment. 
Due to the following observation, in this work, we focus on the first and  second moments. However, our methodology can be extended to the higher moments. 
\begin{lemma}\label{eq:gaussian}
    Assume for some $l$, $\mathscr{X}^{[l]}_{a}$ follows a Multivariate normal distribution for $a\in \{1,2\}$. If  $\mathscr{Z}_{1}^{[l]}$ and  $\mathscr{Z}_{2}^{[l]}$ have the same mean value and covariance matrix, then $\hat{\mathscr{Y}}$ is independent of $\mathscr{A}.$
\end{lemma}

In the rest of this paper, we  assume that we have access to $N = N_1+N_2$ data samples where $N_{1}$ samples belong to the first group and $N_{2}$ samples belong to the second group. We denote the input of layer $l$ associated with these $N$ samples by $X^{[l]} = [X_{1}^{[l]},X_{2}^{[l]}] \in \mathbb{R}^{N \times n^{[l]}}$ where $X_{1}^{[l]} = [(x_{1}^{{[l]}(1)})^{T}, ..., (x_{1}^{{[l]}(N_{1})})^{T}]^T \in \mathbb{R}^{N_{1} \times n^{[l]}}$ are the data points that belong to group  $a_1$ and $X_{2}^{[l]} = [(x_{2}^{{[l]}(1)})^{T}, ..., (x_{2}^{{[l]}(N_{2})})^{T}]^T \in \mathbb{R}^{N_{2} \times n^{[l]}}$ are the data points that belong to group $a_2$.\footnote{We use capital letters for denoting matrices.} 
Using these samples, we will adjust the weight matrix $W^{[l]}$ to achieve better fairness. Since our algorithm for adjusting  $W^{[l]}$ will be applied to a single layer, in the following sections, we omit the superscription $[l]$ when there is no ambiguity.
\section{EXPLAINABLE SVD}
In this part, our goal is to use singular value decomposition to understand contributing factors to unfairness. In particular, we will introduce a linear transformation $S$ for weight matrix $W$ such that the singular values and singular vectors of $WS$ are associated with the disparities between two demographic groups. We will use the SVD of $WS$ to adjust matrix $W$ to improve fairness.    

\subsection{Explainable SVD for First Moment}
Let $\{x_{1}^{(1)},..., x_{1}^{(N_{1})}\}$ and $\{x_{2}^{(1)},..., x_{2}^{(N_{2})}\}$ be the rows of $X_{1}$ and $X_{2}$. Then, the expected values of $\mathscr{X}_1$ and $\mathscr{X}_2$ can be estimated as follows: 
\begin{align}
    \bar{x}_{1} = \frac{1}{N_{1}}\sum_{i=1}^{N_{1}}x_{1}^{(i)} ,~~~~~ \bar{x}_{2} = \frac{1}{N_{2}}\sum_{i=1}^{N_{2}}x_{2}^{(i)}.\nonumber
\end{align}
Consider the linear transformations $\mathscr{X}_1 W^{T}$ and $\mathscr{X}_2 W^{T}$. Then, $\mathbb{E}\{\mathscr{X}_i W^{T}\}$ can also be estimated by   $\bar{x}_{i} W^{T}$. We denote the squared Frobenius distance  between $\bar{x}_1 $ and $\bar{x}_2$ after linear transformations $\bar{x}_1 W^T$ and $\bar{x}_2  W^T$ as $d_{e}^{2}$. That is,
\begin{align}
    d_{e}^{2}(\bar{x}_{1}, \bar{x}_{2}; W) = \left\|\bar{x}_{1}W^{T} - \bar{x}_{2}W^{T}\right\|_{F}^{2}.
\end{align}

\ul{Our goal is to take advantage of SVD and find a new matrix $W'$ that is close to $W$ and for which $d_{e}^{2}(\bar{x}_{1}, \bar{x}_{2}; W')$ is almost zero.} In this case, intuitively, we can make sure the output from the linear transformation defined by $W'$ has the same expected value across different demographic groups. To find $W'$, we first introduce an auxiliary matrix $S_e$ based on the following lemma:
\begin{lemma}\label{lemma1}
   For any $\epsilon_{e}>0$, there exists an invertible matrix $S_{e}$ such that
    \begin{align}\label{eq:lemma_1}
        S_{e}S_{e}^{T} = (\bar{x}_{1} - \bar{x}_{2})^{T}(\bar{x}_{1} - \bar{x}_{2}) + \epsilon_{e} I,
    \end{align}
\end{lemma}
As we will see in the next theorem, $d_e^2(\bar{x}_1,\bar{x_2}; W)$ can be calculated using singular values of $W S_e$. 
\begin{theorem}\label{the1}
    For vectors $\bar{x}_{1}$, $\bar{x}_{2}$ and $W$, if $S_{e}$ satisfies $S_{e}S_{e}^{T} = (\bar{x}_{1} - \bar{x}_{2})^T(\bar{x}_{1} - \bar{x}_{2}) + \epsilon_e I$, and the SVD of $WS_{e}$ is given by $WS_{e} = \sum_{i=1}^{r_{e}}\sigma_{i(e)}u_{i(e)}v_{i(e)}^{T},$
    then,
    \begin{align}\label{eq:w1}
        d_{e}^{2}(\bar{x}_{1}, \bar{x}_{2}; W) = \sum_{i=1}^{r_{e}}\sigma_{i(e)}^{2} - \epsilon_e 
        \mathrm{tr}[WW^T].
    \end{align}
\end{theorem}

Give the above theorem, we have the following corollary:
\begin{corollary}\label{cor1}
    If $S_{e}S_{e}^{T} = (\bar{x}_{1} - \bar{x}_{2})^{T}(\bar{x}_{1} - \bar{x}_{2}) + \epsilon_{e} I$, with the same SVD decomposition in Theorem~\ref{the1}, we have,
    \begin{align}
        d_{e}^{2}(\bar{x}_{1}, \bar{x}_{2}; W) < \sum_{i=1}^{r_{e}}\sigma_{i(e)}^{2} = ||WS_e||^2_F.
    \end{align}
\end{corollary}

Given the above corollary, the difference in the (empirical) mean values of $\mathscr{Z}_1$ and $\mathscr{Z}_2$ can be \textit{explained by the SVD} of $WS_e$ and is bounded by $||WS_e||^2_F$. To decrease the disparity between the two groups, this observation encourages us to replace $W$ with $W'$ such that $||W'S_e||^2_F < ||WS_e||^2_F$. In Section \ref{sec:OPT}, we take advantage of the SVD of $WS_e$ and propose an \textit{efficient} method for finding $W'$ that improves fairness. 


\subsection{Explainable SVD for Second Moment}
Define $\bar{X}_{1} \in \mathbb{R}^{N_{1} \times n}$, each row of $\bar{X}_{1}$ is a copy of $\bar{x}_{1}$. $\bar{X}_{2}\in \mathbb{R}^{N_{2} \times n}$, each row of $\bar{X}_{2}$ is a copy of $\bar{x}_{2}$. Then, unbiased estimates for the covariance matrices (i.e., the second moments)  $Var(\mathscr{X}_{1})$ and $Var(\mathscr{X}_{2})$ \citep{rohatgi2015introduction} are given by:   
\begin{align}
    Var(X_{1}) = \frac{1}{N_{1} - 1}(X_{1} - \bar{X}_{1})^{T}(X_{1} - \bar{X}_{1}), \nonumber\\
    Var(X_{2}) = \frac{1}{N_{2} - 1}(X_{2} - \bar{X}_{2})^{T}(X_{2} - \bar{X}_{2}). \nonumber
\end{align}
After applying the linear transformation $\mathscr{Z}_a = \mathscr{X}_a W^T$, the covariance matrix of $\mathscr{Z}_a$ can be stimated by  $\frac{1}{N_{a} - 1}W(X_{a} - \bar{X}_{a})^{T}(X_{a} - \bar{X}_{a})W^{T}, a\in \{1,2\}$. 
 Denote $\tilde{X}_{1} = X_{1} - \bar{X}_{1}$ and $\tilde{X}_{2} = X_{2} - \bar{X}_{2}$, the squared Frobenius distance $d_{v}^{2}$ between the empirical covariance matrices of $\mathscr{Z}_1$ and $\mathscr{Z}_2$ is given by: 
{
\begin{align}
    d_{v}^{2}(X_{1}, X_{2}; W) = \left\|\frac{W\tilde{X}_{1}^{T}\tilde{X}_{1}W^{T}}{N_{1} - 1} - \frac{W\tilde{X}_{2}^{T}\tilde{X}_{2}W^{T}}{N_{2} - 1}\right\|_{F}^{2}.
\end{align}
}
Therefore,  $d_{v}^{2}(X_{1}, X_{2}; W) = \left\|WMW^{T}\right\|_{F}^{2}$, where 
\begin{align}\label{eq:defM}
    M = \frac{1}{N_{1} - 1}\tilde{X}_{1}^{T}\tilde{X}_{1} - \frac{1}{N_{1} - 1}\tilde{X}_{2}^{T}\tilde{X}_{2}.
\end{align}

Since matrix $M$ is symmetric, its eigenvalue decomposition is given by $Q\Lambda Q^{T}$ where $Q$ is an orthogonal matrix and  $\Lambda =\mathrm{diag}([\lambda_1,\lambda_2,\ldots,\lambda_n]) $  is a diagonal matrix.  

Let $|\Lambda| =\mathrm{diag}(\left[|\lambda_1|,|\lambda_2|,\ldots,|\lambda_n|\right])$, $|\Lambda|^{\frac{1}{2}} =\mathrm{diag}([\sqrt{|\lambda_1|},\sqrt{|\lambda_2|},\ldots,\sqrt{|\lambda_n|}])$, 
$|M| = Q|\Lambda|Q^{T}$, and $S_{v} = Q\left|\Lambda\right|^{\frac{1}{2}}$. Note that $S_{v}S_{v}^{T} = \left|M\right|$. 
We have the following theorem for $d_{v}^{2}(X_{1}, X_{2}; W)$.

\begin{theorem}\label{the2}
    For any matrix $X_{1}, X_{2}$ and $W$, if $S_{v}$ satisfies $S_{v}S_{v}^{T} = \left|M\right|$, and the SVD decomposition of $WS_{v}$ is given by
$   
     WS_{v} = \sum_{i=1}^{r_{v}}\sigma_{i(v)}u_{i(v)}v_{i(v)}^{T},
    $
    then,
    \begin{align}\label{eq:w2}
        d_{v}^{2}(X_{1}; X_{2}; W) \leq 
        ||WS_v||^4_{4}  = \sum_{i=1}^{r_{v}}\sigma_{i(v)}^{4},
    \end{align}
where $||.||_{4}$ is the Schatten-4 norm. 
\end{theorem}
According to Theorem~\ref{the2}, the difference in the (empirical) second moments of $\mathscr{Z}_1$ and $\mathscr{Z}_2$ can be \textit{explained by the SVD} of $WS_v$ and is bounded by Schatten-4 norm of $WS_v$. To decrease the disparity among the two groups, this observation encourages the replacement of $W$ with $W'$ such that $||W'S_v||_{4}^{4} < ||WS_v||_{4}^{4}$. In Section \ref{sec:OPT}, we leverage the SVD of $WS_v$ to find a $W'$ that leads to outputs with equivalent second moments across different demographic groups.

\section{EXPLAINABLE SVD ALGORITHM}\label{sec:OPT}
We describe how to obtain a fair weight matrix $W^{*}$ through the explainable SVD. According to Theorem~\ref{the1} and the accompanying corollary, we can consistently restrict the upper bound of $d_{e}^{2}$ by constraining the squared summation of $\sigma_{i(e)}$. When $\epsilon_{e}$ is 0, it is fully explainable as the summation precisely predicts $d_{e}^{2}$ accurately. Even when $\epsilon_{e}$ is not 0, given that it can always be a very small number, the upper bound in Corollary~\ref{cor1} generally remains very tight.

From Theorem~\ref{the2}, we know that the upper bound of $d_{v}^{2}$ can be restricted by adding a constraint on the fourth power summation of the singular values. Similar to the first moment case, when $M$ is positive definite, we have that the upper bound is precisely equal to $d_{v}^{2}(X_{1}, X_{2}; W)$. 
This allows us to understand the reduction in $d_{v}^{2}$ using only the information from $\sigma_{i(v)}$.
\subsection{Optimization}
In this part, our goal is to propose an optimization problem to replace the matrix $W$ in the regression model $f$ with a matrix $W'$ such that $X_1W^{'T}$ and $X_2W^{'T}$ have the same first and second moments. 


To ensure that $X_1W^{'T}$ and $X_2W^{'T}$ share the same first moment, we consider the following optimization problem,
\begin{align}\label{eq:opt1}
    \min_{W^{'}} \left\|XW^{T} - XW^{'T}\right\|_{F}^{2}  s.t., d_{e}^{2}(\bar{x}_{1}, \bar{x}_{2}; W') \leq c_{e},
\end{align}
where $c_e$ is a constant (hyper-parameter). 
The above objective function implies that $W'$ should have the same performance as $W$, and the constraint ensures that  $X_1W^{'T}$ and $X_2W^{'T}$ have the same first moment. While the above optimization problem is convex, it does not have a closed-form solution. It can be computationally expensive to solve and requires a polynomial-time algorithm \citep{tseng1988simple}. 

To make the optimization problem \eqref{eq:opt1} tractable, we limit the feasible set to the following,
\begin{align}
    \mathcal{W}_e = \left\lbrace W': W' = \left(\sum_{i=1}^{r_{e}}\sigma_{i(e)}^{'}u_{i(e)}v_{i(e)}^{T}\right)S_{e}^{-1} \right\rbrace,
\end{align}
Note that if $W' = \left(\sum_{i=1}^{r_{e}}\sigma_{i(e)}^{'}u_{i(e)}v_{i(e)}^{T}\right)S_{e}^{-1}$, then by Corollary \ref{cor1}$, 
    d_{e}^{2}(\bar{x}_{1}, \bar{x}_{2}; W') \leq  \sum_{i=1}^{r_{e}}\sigma_{i(e)}^{'2}$. 
Considering $\mathcal{W}_e$ as the feasible set for \eqref{eq:opt1}, the optimization problem can be reformulated as follows, 
{\small
\begin{align}\label{optimization1}
    & \sigma_{i(e)}^{*} = \argmin_{\sigma_{i(e)}^{'}, i\leq r_e}\left\|X(S_{e}^{-1})^{T}\left(\sum_{i=1}^{r_{e}}(\sigma_{i(e)}^{'} - \sigma_{i(e)} )u_{i(e)}v_{i(e)}^{T} \right)\right\|_{F}^{2} \nonumber \\
    & s.t. \sum_{i=1}^{r_{e}} \sigma_{i(e)}^{'2} \leq c_{e}.
\end{align}}This optimization problem has a closed-form solution, detailed in the following theorem.
\begin{theorem}\label{the:opt1}
    The solution to optimization problem~\eqref{optimization1} is given by,
    \begin{align}
        \sigma_{i(e)}^{*} = \frac{\sigma_{i(e)}k_{i(e)}}{k_{i(e)}\gamma_{e}},
    \end{align}
    where $k_{i(e)} = v_{i(e)}^{T}(S_{e}^{-1})X^{T}X(S_{e}^{-1})^{T}v_{i(e)}$, and $\gamma_{e}$ is the solution of the equation
    \begin{align}\label{eq:lam_1_solution}
        \sum_{i=1}^{r_{e}}\left(\frac{\sigma_{i(e)}k_{i(e)}}{k_{i(e)}\gamma_{e}}\right)^{2} = c_{e}.
    \end{align}
\end{theorem}
Equation~\ref{eq:lam_1_solution} is a single variable equation of $\gamma_{e}$. Therefore, the value of $\gamma_{e}$ can be obtained by numerical methods such as Newton-Raphson method \citep{ypma1995historical}. After solving optimization problem \eqref{optimization1}, we create  matrix $W_e^{*} = \left(\sum_{i=1}^{r_{e}}\sigma_{i(e)}^{*}u_{i(e)}v_{i(e)}^{T}\right)S_{e}^{-1}$ and replace $W$ with $W_e^{*}$.

Similarly, to ensure that $X_1W^{'T}$ and $X_2W^{'T}$ have the same second moment, we consider the following optimization problem,
\begin{align}\label{eq:original_opt_2}
    \min_{W'} \left\|{XW^{T} - XW^{'T}}\right\|_{F}^{2} 
    s.t., d_{v}^{2}(X_{1}, X_{2}; W') \leq c_{v}.
\end{align}
To make the above optimization problem tractable, we consider the following feasible set\footnote{When $M$ is not full rank, $S_{v}^{-1}$ can be replaced by the pseudo-inverse of $S_{v}$ \citep{bjerhammar1951application}.}, \begin{align}
    \mathcal{W}_v = \left\lbrace W': W' = \left(\sum_{i=1}^{r_{v}}\sigma_{i(v)}^{'}u_{i(v)}v_{i(v)}^{T}\right)S_{v}^{-1} \right\rbrace,
\end{align}
By Theorem~\ref{the2}, if $W' = \left(\sum_{i=1}^{r_{v}}\sigma_{i(v)}^{'}u_{i(v)}v_{i(v)}^{T}\right)S_{v}^{-1}$,  then
    $d_{v}^{2}(X_{1}, X_{2}; W') \leq  \sum_{i=1}^{r_{v}}\sigma_{i(v)}^{'4}$. By restricting the feasible set to $\mathcal{W}_v$, we can re-write the optimization problem as follows,
{\small
\begin{align}\label{optimzation2}
    & \sigma_{i(v)}^{*} = \argmin_{\sigma_{i(v)}^{'}, i\leq r_{v}}\left\|X(S_{v}^{-1})^{T}\left(\sum_{i=1}^{r_{v}}(\sigma_{i(v)}^{'} - \sigma_{i(v)})u_{i(v)}v_{i(v)}^{T} \right)\right\|_{F}^{2} \nonumber \\
    & s.t., \sum_{i=1}^{r_{v}} \sigma_{i(v)}^{'4} \leq c_{v}.
\end{align}
}The above optimization problem is  convex and the solution is given by the following theorem:
\begin{theorem}\label{the:opt2}
   The following is the solution to optimization problem \eqref{optimzation2}, 
    \begin{align}\label{eq:sigma2}
        \sigma_{i(v)}^{*} =  -\frac{k_{i(v)}}{\phi}    + \frac{\phi}{6^{\frac{2}{3}}\gamma_{v}},
    \end{align}
    
    where $$\phi = 6^{\frac{1}{3}}\left(9\gamma_{v}^{2}k_{i(v)}\sigma_{i(v)} + \sqrt{3}\sqrt{2\gamma_{v}^{3}k_{i(v)}^{3} + 27\gamma_{v}^{4}k_{i(v)}\sigma_{i(v)}^{2}}\right)^{\frac{1}{3}},$$ $$k_{i(v)} = v_{i(v)}^{T}(S_{v}^{-1})X^{T}X(S_{v}^{-1})^{T}v_{i(v)},$$ and $\gamma_{v}$ is the solution of the following equation, 
    \begin{align}\label{eq:lambda2}
        \sum_{i=1}^{r_{v}} \{-\frac{k_{i(v)}}{\phi} + \frac{\phi}{6^{\frac{2}{3}}\gamma_{v}}\}^{4} = c_{v}.
    \end{align}
\end{theorem}
Note that Eq.~\ref{eq:lambda2} can be solved with numerical methods. 

After solving optimization problem \eqref{optimzation2}, we create the matrix $W_v^{*} = \left(\sum_{i=1}^{r_{v}}\sigma_{i(v)}^{*}u_{i(v)}v_{i(v)}^{T}\right)S_{v}^{-1}$ and replace $W$ with $W_v^{*}$. 
In the next part, we take advantage of optimization problems \eqref{optimization1} and \eqref{optimzation2} to ensure both the first and second moments of $X_1W'^T$ and $X_2W'^T$ are the same. 

\subsection{Algorithm} 
\begin{algorithm}[h]
    \caption{Explainable SVD Fairness Enhancement Algorithm (ESVDFair)}\label{algorithm}
    \textbf{Input:} Neural network $f$ with weight matrix $\{W^{[1]}, W^{[2]}, ..., W^{[L]}\}$, layer index $l$, training data $X = [X_{1};X_{2}]$, constants $c_{e}$, $c_{v}$, $\epsilon_{e}$.
    \begin{algorithmic}[1]
        \State Feed $X$, $X_{1}$, $X_{2}$ to the neural network, get the input matrix of layer $l$ as $X^{[l]}$, $X_{1}^{[l]}$ and $X_{2}^{[l]}$.
        \State Compute the average inputs $\bar{x}_{1}^{[l]}$ and $\bar{x}_{2}^{[l]}$. 
        \State $\tilde{X}^{[l]}_{1} \leftarrow X^{[l]}_{1} - \bar{X}^{[l]}_{1}$, $\tilde{X}^{[l]}_{2} \leftarrow X^{[l]}_{2} - \bar{X}^{[l]}_{2}$.
        \State $\{Q,\lambda\} \leftarrow$ spectrum decomposition of $W^{[l]}$.
        \State $\left|\Lambda \right|^{\frac{1}{2}} \leftarrow \mathbf{diag}(|\lambda_{1}|, ..., |\lambda_{n}|)$.
        \State $S_{v} \leftarrow Q\left|\Lambda \right|^{\frac{1}{2}}$.
        \State $\{u_{i(v)}, \sigma_{i(v)}, v_{i(v)}\}_{i=1}^{r_{v}} \leftarrow$ SVD($W^{[l]}S_{v}$).
        \For{$i \leftarrow$ 1 to $r_{v}$}
            \State $k_{i(v)} \leftarrow v_{i(v)}^{T}(S_{v}^{-1})X^{T}X(S_{v}^{-1})^{T}v_{i(v)}$.
        \EndFor
        \State $\gamma_{v} \leftarrow $ solution of Eq.~\ref{eq:lambda2}.
        \For{$i \leftarrow$ 1 to $r_{v}$}
            \State Set $\sigma_{i(v)}^{'}$ as Eq.~\ref{eq:sigma2}.
        \EndFor
        \State  $W_{v}^{*[l]} \leftarrow \left(\sum_{i=1}^{r_{v}}\sigma_{i(v)}^{'}u_{i(v)}v_{i(v)}^{T}\right)S_{v}^{-1}$.
        \State Find $S_{e}$ where  $S_{e}S_e^T =  (\bar{x}_{1}^{[l]} - \bar{x}_{2}^{[l]})^{T}(\bar{x}_{1}^{[l]} - \bar{x}_{2}^{[l]}) + \epsilon_e I$.
        \State $\{u_{i(e)}, \sigma_{i(e)}, v_{i(e)}\}_{i=1}^{r_{e}} \leftarrow$ SVD($W_{v}^{*[l]}S_{e}$).
        \For{$i \leftarrow$ 1 to $r_{e}$}
            \State $k_{i(e)} \leftarrow v_{i(e)}^{T}(S_{e}^{-1})X^{T}X(S_{e}^{-1})^{T}v_{i(e)}$.
        \EndFor
        \State $\gamma_{e} \leftarrow$ \textbf{Solve}$\left(\sum_{i=1}^{r_{e}}\left(\frac{\sigma_{i(e)}k_{i(e)}}{k_{i(e)}\gamma_{e}}\right)^{2} = c_{e}\right)$.
         $\newline$\textbf{Solve} here is an arbitrary numerical method to solve a single variable equation.
        \For{$i \leftarrow$ 1 to $r_{e}$}
            \State $\sigma^{'}_{i(e)} \leftarrow \frac{\sigma_{i(e)}k_{i(e)}}{k_{i(e)}\gamma_{e}}$.
        \EndFor
        \State $W_{e}^{*[l]} \leftarrow \left(\sum_{i=1}^{r_{e}}\sigma_{i(e)}^{'}u_{i(e)}v_{i(e)}^{T}\right)S_{e}^{-1}$.
        \State $W^{[l]} \leftarrow W_{e}^{*[l]}$.
        \State \Return f.
    \end{algorithmic}
\end{algorithm}

Based on the solutions of the two optimization problems \eqref{optimization1} and \eqref{optimzation2}, Algorithm~\ref{algorithm} provides a method for adjusting the weight matrix of an arbitrary  layer $l$ to improve fairness. In particular, this algorithm first changes $W^{[l]}$ to $W_v^{*[l]}$  to equalize the covariance matrices of $\mathscr{Z}_1^{[l]}$ and $\mathscr{Z}_2^{[l]}$ and decreases the  disparity across the two groups in terms of covariance matrices. Then, we replace $W$ in \eqref{eq:opt1} by  $W_v^{*[l]}$ and solve the optimization problem \eqref{optimization1} to get $W_{e}^{*[l]}$. Finally, we replace the original weight matrix $W^{[l]}$ by $W_{e}^{*[l]}$. Based on Lemma \ref{eq:gaussian}, we expect this procedure mitigate the disparity and improve fairness in terms of statistical parity. 

 To enhance accuracy, we update the weight matrix of the last layer using the ordinary least square problem, 
\begin{align}
    W^{*[L]} &=& \argmin_{W^{[L]}} \left\|X^{[L]}W^{[L]} - Y\right\|_{F}^{2}\nonumber  \\ &=& \left(X^{[L]T}X^{[L]}\right)^{-1}X^{[L]T}Y.
\end{align}
Note that Lemma \ref{eq:gaussian} holds even after updating the weight matrix of the last layer. 
In Algorithm~\ref{algorithm2}, we combine the least square algorithm with Algorithm \ref{algorithm}, to ensure both fairness improvement and high accuracy. 
\begin{algorithm}[h]
    \caption{ESVDFair Algorithm with Adjustment}\label{algorithm2}
    \textbf{Input:} Neural network $f$ with weight matrix $\{W^{[1]}, W^{[2]}, ..., W^{[L]}\}$, training data $X = [X_{1};X_{2}]$, $Y$, constants $c_{e}$, $c_{v}$, $\epsilon_{e}$.
    \begin{algorithmic}[1]
        \State $W^{[L - 1]} \leftarrow$ ESVDFair($f$, $L-1$, $X = [X_{1}; X_{2}]$, $c_{e}$, $c_{v}$, $\epsilon_{e}$). 
        \State $W^{[L]} \leftarrow \left(X^{[L]T}X^{[L]}\right)^{-1}X^{[L]T}Y$.
        \State \Return f.
    \end{algorithmic}
\end{algorithm}
Note that in Algorithm \ref{algorithm2}, we apply the  ESVDFair algorithm to the second-last layer of the neural network. However, ESVDFair can be applied to any other layer. We also want to emphasize that in Algorithm \ref{algorithm2}, we only adjust the weights of two layers. 
\section{EXPERIMENT}\label{sec:exp}
In this section, we conduct empirical studies for our proposed algorithm on two real datasets.
\begin{figure*}[t]
    \centering
    \begin{subfigure}[b]{0.245\textwidth}
        \includegraphics[width=\textwidth]{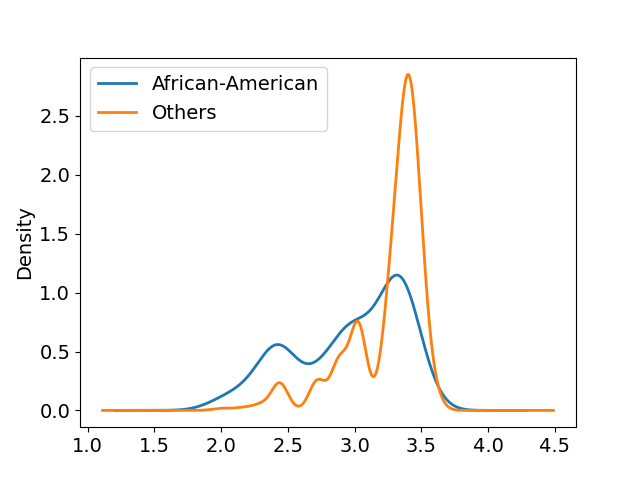}
        \caption{No post-processing}\label{fig:law_distribution_original}
    \end{subfigure}
    \begin{subfigure}[b]{0.245\textwidth}
        \includegraphics[width=\textwidth]{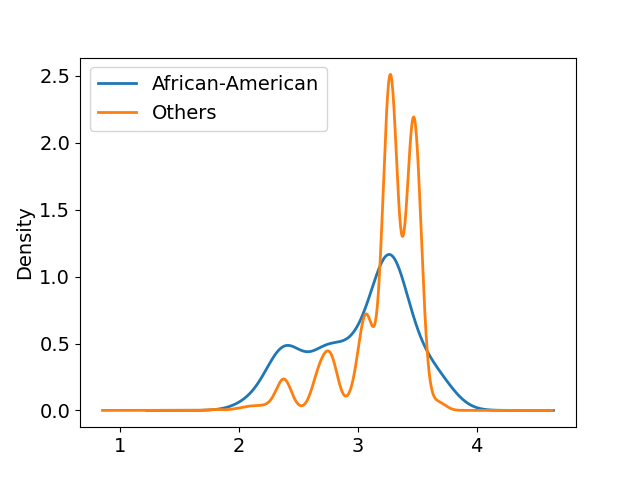}
        \caption{\cite{chzhen2020fair}}\label{fig:law_distribution_chzhen}
    \end{subfigure}
    \begin{subfigure}[b]{0.245\textwidth}
        \includegraphics[width=\textwidth]{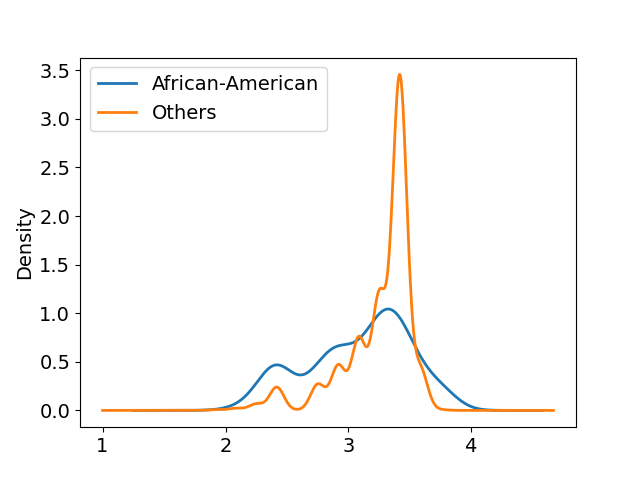}
        \caption{\cite{xian2024differentially}}\label{fig:law_distribution_private}
    \end{subfigure}
    \begin{subfigure}[b]{0.245\textwidth}
        \includegraphics[width=\textwidth]{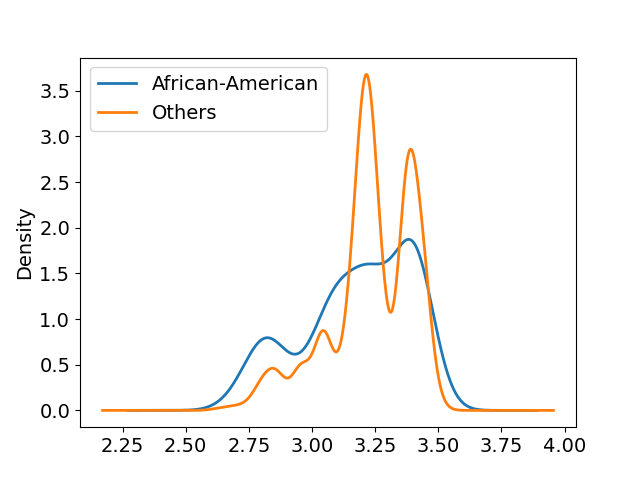}
        \caption{ours}\label{fig:law_distribution_ours}
    \end{subfigure}
    \caption{Density of the output distribution across two sensitive groups on the Law School Sucess dataset.}\label{fig:law_distribution}
\end{figure*}
\subsection{Datasets}
We use the Law School Success dataset \citep{wightman1998lsac} and the COMPAS dataset \citep{washington2018argue} to evaluate our proposed method. 

The Law School Success dataset contains 22,407 pieces of student record. We follow the experiment in \cite{xian2024differentially} and use 6 attributes in the dataset: dnn bar pass prediction (the LSAT prediction from a DNN model, ranging from 0 to 1), gender (gender of the student, which could be male or female), lsat (LSAT score received by the student, ranging from 0 to 1), race (Black or White), pass bar (whether or not the student eventually pass the bar) and ugpa (student's undergraduate GPA ranging from 0 to 4). In this experiment, we focus on two racial groups: Black and White. We choose race as the sensitive attribute and  ugpa as the target attribute and the remaining attributes as features. 

The COMPAS dataset is a collection of 11,001 records that track the recidivism rates of individuals convicted of crimes. We divide the dataset into two groups based on the sensitive attribute of race: one group consists of African Americans, while the other group includes individuals of all other racial backgrounds. Our goal is to predict whether a person will commit another crime within the next two years, using various features such as age, sex, and type of assessment.

\subsection{Baselines}
Since our method belongs to the post-processing methods for fair regression, we use the state-of-the-art post-processing methods proposed by \cite{chzhen2020fair} and \cite{xian2024differentially} as our baselines.

\cite{chzhen2020fair} links the optimal fair predictor problem to a Wasserstein barycenter problem \citep{agueh2011barycenters}. Given a pre-trained unfair predictor $f$, the fair predictor is given by,
\begin{align}\label{eq:baseline1}
    g(x, a_1) &= \Pr\{\mathscr{A} = a_2\}\cdot \mathcal{Q}_{\hat{\mathscr{Y}}|\mathscr{A}=a_2} \circ \mathcal{F}_{\hat{\mathscr{Y}}|\mathscr{A}=a_1}(f(x)),\nonumber\\
    a_1&\in \{1,2\}, a_2\in \{1,2\}, a_1\neq a_2, 
\end{align}
where $\mathcal{Q}_{\hat{\mathscr{Y}}|\mathscr{A}=a_2}(.) $  is the quantile function of the conditional distribution $\hat{\mathscr{Y}}|\mathscr{A}=a_2$, and $\mathcal{F}_{\hat{\mathscr{Y}}|\mathscr{A}=a_1}(.)$ is the cumulative distribution function of the conditional distribution $\hat{\mathscr{Y}}|\mathscr{A}=a_1$. 

\cite{xian2024differentially} use a similar idea of computing Wasserstein barycenter. Their method is composed of three steps, estimating the output distributions, computing the Wasserstein barycenter and finding the optimal transports to the barycenter. Given an unfair predictor $f$, the fair predictor is given by,
\begin{align}\label{eq:baseline2}
    g(x, a) = t_{a} \circ h \circ f(x),
\end{align}
where $h$ is a discretizer over $\mathcal{Y}$, and $t_{a}: \mathcal{Y} \rightarrow \mathcal{Y}$ is a function that represents the optimal transports.

\begin{figure*}[t]
    \centering
    \begin{subfigure}[b]{0.24\textwidth}
        \includegraphics[width=\textwidth]{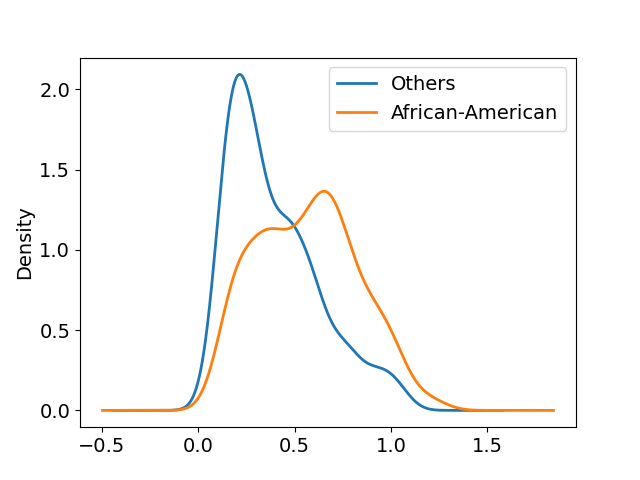}
        \caption{No post-processing}\label{fig:compas_distribution_original}
    \end{subfigure}
    \begin{subfigure}[b]{0.24\textwidth}
        \includegraphics[width=\textwidth]{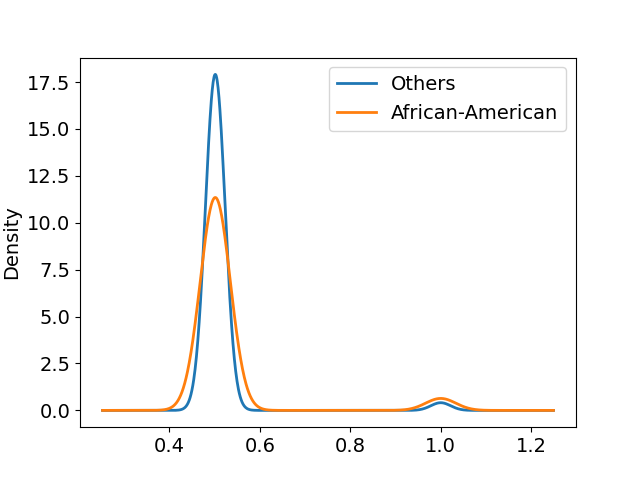}
        \caption{\cite{chzhen2020fair}}\label{fig:compas_distribution_chzhen}
    \end{subfigure}
    \begin{subfigure}[b]{0.24\textwidth}
        \includegraphics[width=\textwidth]{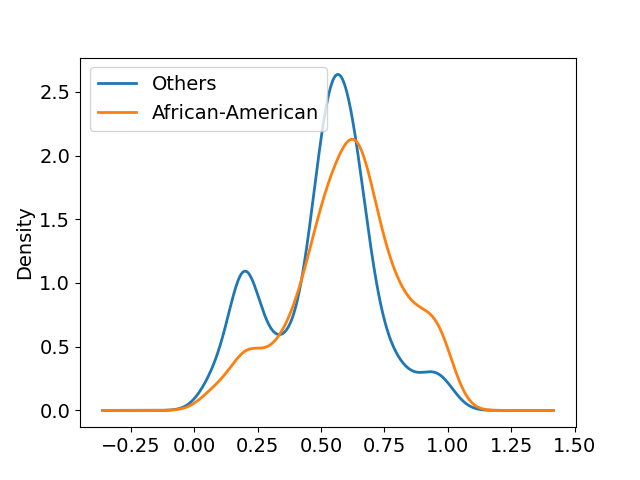}
        \caption{\cite{xian2024differentially}}\label{fig:compas_distribution_private}
    \end{subfigure}
    \begin{subfigure}[b]{0.24\textwidth}
        \includegraphics[width=\textwidth]{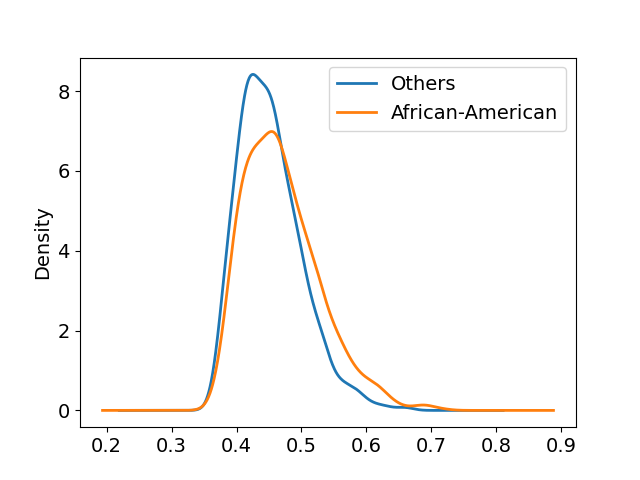}
        \caption{ours}\label{fig:compas_distribution_ours}
    \end{subfigure}
    \caption{Density of the output distribution across two sensitive groups on the COMPAS dataset.}\label{fig:compas_distribution}
\end{figure*}
\subsection{Implementations}
In the experiments, we split the dataset into a training dataset, a validation dataset and a test dataset with a ratio of $70\% \sim 15\% \sim 15\%$ randomly for 50 times. The experiments are repeated 50 times to evaluate the average performance. We use a five layer neural network as the unfair predictor. The architecture of the neural network is displayed in Table~\ref{tab:nn_architecture} in the appendix.

For each experiment, we pre-train the neural network on the training data for 20 epochs. The optimizer is the Adam optimizer with an initial learning rate 1e-3. After each epoch, the learning rate decays by a ratio of 0.8. We assume that at the inference time, we do not have access to the sensitive attribute $\mathscr{A}$ and the neural network does not accept $\mathscr{A}$ as an input feature. However, the sensitive attributes associated with samples $X$ is available for running  Algorithm \ref{algorithm2} and for  estimating $t_a$, $\mathcal{Q}_{\hat{\mathscr{Y}}|\mathscr{A}}$, and $\mathcal{F}_{\hat{\mathscr{Y}}|\mathscr{A}}$ in baselines. Note that after adjusting the regression model $f$, our method does not need the sensitive attribute during the inference time. On the other hand, we can see in \eqref{eq:baseline1} and \eqref{eq:baseline2}, the  baselines need the sensitive attribute at the inference time. As a result,  we train a logistic regression model to predict sensitive attributes and then use the predicted sensitive attributes at the test time for the law school dataset. For the COMPAS dataset, because logistic regression model cannot converge on the training data, we train a model with the same architecture in Table~\ref{tab:nn_architecture} (except for the output layer which uses sigmoid activation function).

For our method, we set $\epsilon_{e}$ to 1e-5. Using the validation dataset, we tune hyper-parameters $c_{e}$ and $c_{v}$ on the Law School Success dataset and set $c_{e} = \frac{1}{\tilde{c}_{e}}\sum_{i=1}^{r_{e}}\sigma_{i(e)}^{2}$ where $\tilde{c}_{e}$ is 15 and $c_{v} = \frac{1}{\tilde{c}_{v}}\sum_{i=1}^{r_{v}}\sigma_{i(v)}^{4}$ where $\tilde{c}_{v}$ is 150. Instead of using Algorithm \ref{algorithm2}, we use gradient descent to fine tune $W^{[5]}$ for 50 epochs after adjusting $W^{[4]}$ using Algorithm~\ref{algorithm2}.
We use MSE and KS as evaluation metrics. MSE is the mean squared error. KS is the empirical version of Eq.~\ref{eq:ks} for measuring fairness. A smaller KS implies a better level of fairness.
\subsection{Results and Analysis}
\begin{table}[htbp]
    \centering
    \caption{The experiment results on the law school dataset for our method and baselines. We report MSE and KS  between two sensitive groups.}\label{tab:results_law}
    \begin{tabular}{ccc}
    \toprule
    Method & MSE & KS \\
    \midrule
    No post-processing & 0.055 $\pm$ 0.006 & 0.376 $\pm$ 0.031 \\
    \cite{chzhen2020fair} & 0.061 $\pm$ 0.007 & 0.267 $\pm$ 0.034 \\
    \cite{xian2024differentially} & 0.062 $\pm $ 0.006 & 0.258 $\pm$ 0.042 \\
    ESVDFair & 0.088 $\pm$ 0.007 & 0.235 $\pm$ 0.042 \\
    \bottomrule
    \end{tabular}
\end{table}
Table~\ref{tab:results_law} shows the results on the Law School Success dataset. For this dataset, we split the range of $Y$ into 36 bins for the baselines. We can see that our ESVDFair algorithm achieves a similar fairness-accuracy trade-off with the baselines. Note that, in contrast to the baselines, our method can improve fairness without using the sensitive attribute at the inference time.
Figure~\ref{fig:law_distribution} visualizes the output distribution of the model modified by our method and the two baselines; this suggests that our method can effectively align the two distributions. The baselines also manage to align the distributions, a finding consistent with Table~\ref{tab:results_law} where MSE and KS values are the similar for our method and the baselines.

\begin{table}[htbp]
    \centering
    \caption{The experiment results on COMPAS dataset for our method and baselines. We report MSE and KS  between two sensitive groups.}\label{tab:results_compas}
    \begin{tabular}{ccc}
    \toprule
    Method & MSE & KS \\
    \midrule
    No post-processing & 0.185 $\pm$ 0.006 & 0.256 $\pm$ 0.030 \\
    \cite{chzhen2020fair} & 0.403 $\pm$ 0.147 & 0.271 $\pm$ 0.025 \\
    \cite{xian2024differentially} & 0.227 $\pm $ 0.021 & 0.156 $\pm$ 0.053 \\
     ESVDFair & 0.214 $\pm$ 0.004 & 0.130 $\pm$ 0.035 \\
    \bottomrule
    \end{tabular}
\end{table}

\begin{figure*}[t]
    \centering
    \begin{subfigure}[b]{0.19\textwidth}
        \includegraphics[width=\textwidth]{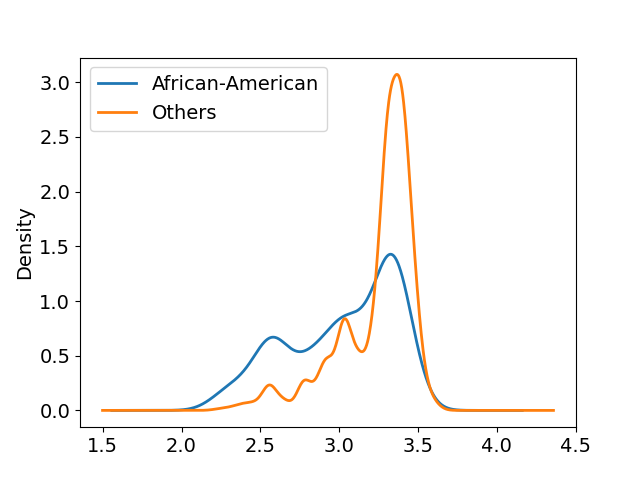}
        \caption{$\tilde{c}_{e} = 1.5$}\label{fig:law_distribution_c_1_1.5}
    \end{subfigure}
    \begin{subfigure}[b]{0.19\textwidth}
        \includegraphics[width=\textwidth]{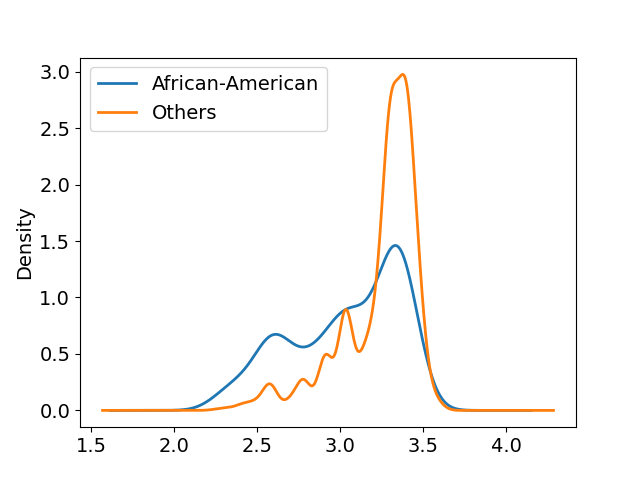}
        \caption{$\tilde{c}_{e} = 2$}\label{fig:law_distribution_c_1_2}
    \end{subfigure}
    \begin{subfigure}[b]{0.19\textwidth}
        \includegraphics[width=\textwidth]{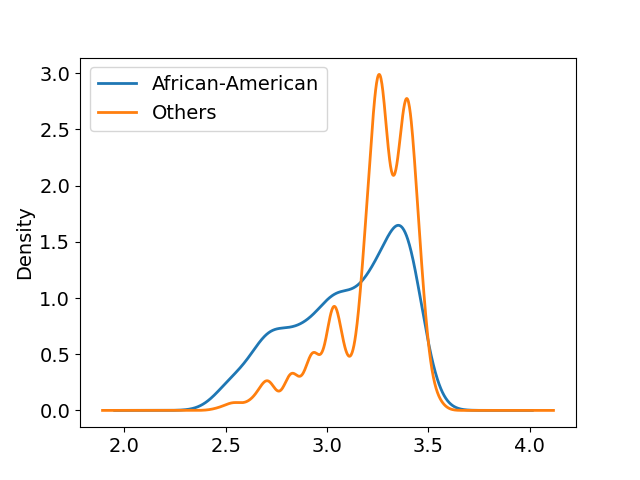}
        \caption{$\tilde{c}_{e} = 5$}\label{fig:law_distribution_c_1_5}
    \end{subfigure}
    \begin{subfigure}[b]{0.19\textwidth}
        \includegraphics[width=\textwidth]{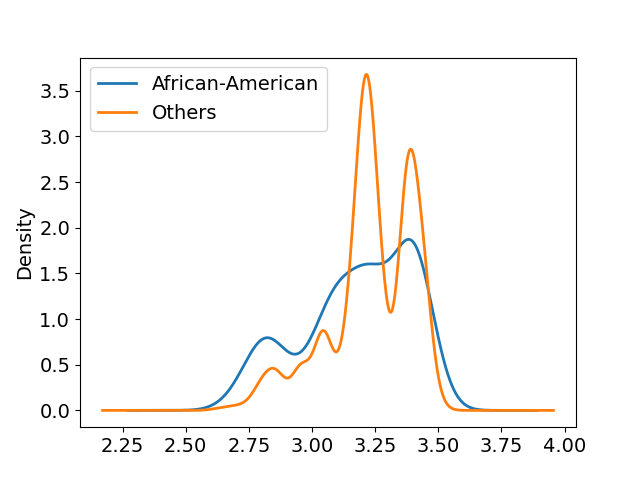}
        \caption{$\tilde{c}_{e} = 15$}\label{fig:law_distribution_c_1_15}
    \end{subfigure}
    \begin{subfigure}[b]{0.19\textwidth}
        \includegraphics[width=\textwidth]{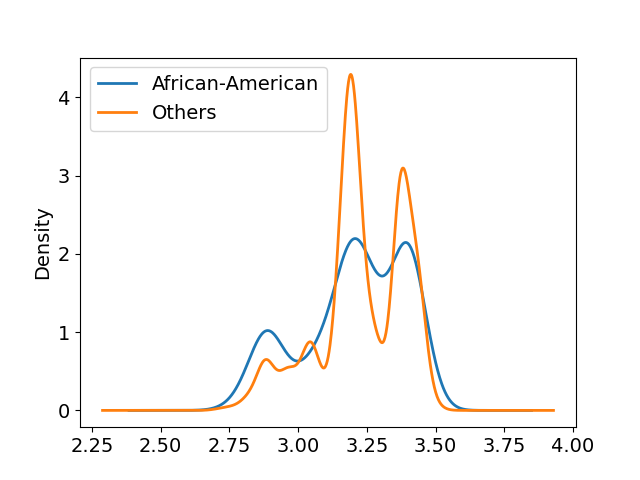}
        \caption{$\tilde{c}_{e} = 50$}\label{fig:law_distribution_c_1_50}
    \end{subfigure}
    \caption{Density of the output distribution across two sensitive groups on the Law School Success dataset with different $\tilde{c}_{e}$.}\label{fig:law_distribution_c_1}
\end{figure*}

We repeat the experiment for the COMPAS dataset and report the results   in Table~\ref{tab:results_compas}. For this dataset, we split the range of $Y$ into 18 bins for the baselines. Table~\ref{tab:results_compas} demonstrates our method can simultaneously impprove both MSE and KS compared to the baselines, achieving a better fairness-accuracy trade-off.  
Figure~\ref{fig:compas_distribution} displays that the distribution of the model's output across the two groups after applying our method and the baselines. The figure illustrates that our method is the most effective at aligning the two distributions.
\begin{figure}
    \centering    \includegraphics[width=\linewidth]{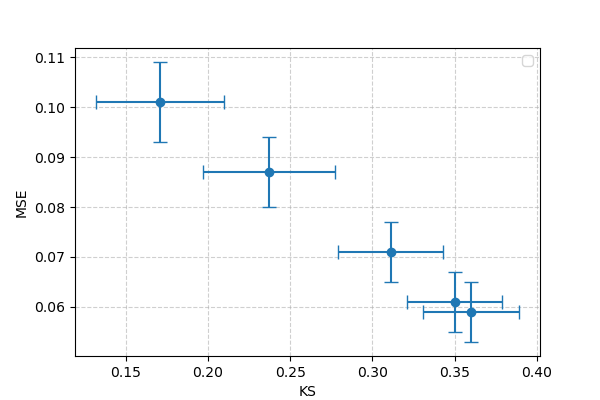}
    \caption{KS vs. MSE with different $\tilde{c}_{e}$}
    \label{fig:c_1}
\end{figure}

\subsection{Effect of $c_{e}$ and $c_{v}$}
In this section, we study the impact of  $c_{e}$ and $c_{v}$ on the distribution of the model's output after post-processing.

In the first step, we set $\tilde{c}_{v}$ as 150 and $\tilde{c}_{e} \in \{1.5, 2, 5, 15, 50\}$ to adjust $W^{[4]}$. Then we fine tune $W^{[5]}$.
Figure~\ref{fig:c_1} illustrates the MSE and KS for different values of $\tilde{c}_{e}$. We observe that with larger $\tilde{c}_{e}$ (smaller $c_{e}$), we achieve better fairness in terms of KS but worse accuracy in terms of MSE. Figure~\ref{fig:law_distribution_c_1} shows the output distribution across the two groups; as $\tilde{c}_{e}$ increases (and $c_e$
  decreases), the output distribution becomes more aligned across different groups, indicating a reduction in disparity.

To validate the effectiveness of $\tilde{c}_{v}$, we fix $c_{e}$ as 15 and utilize $\tilde{c}_{v} \in \{5, 10, 50, 100, 150\}$ in Table~\ref{tab:c_2}.
\begin{table}[htbp]
    \centering
    \caption{The MSE and KS on the Law School dataset for ESVDFair algorithm with different $\tilde{c}_{v}$.}\label{tab:c_2}
    \begin{tabular}{ccc}
    \toprule
    $\tilde{c}_{v}$ & MSE & KS \\
    \midrule
    5 & 0.067 $\pm$ 0.006 & 0.277 $\pm$ 0.041 \\
    10 & 0.070 $\pm$ 0.006 & 0.267 $\pm$ 0.040 \\
    50 & 0.079 $\pm $ 0.007 & 0.246 $\pm$ 0.041 \\
    100 & 0.085 $\pm$ 0.007 & 0.241 $\pm$ 0.041 \\
    150 & 0.088 $\pm$ 0.007 & 0.237 $\pm$ 0.040 \\
    \bottomrule
    \end{tabular}
\end{table}
With larger $\tilde{c}_{v}$ (resulting in a smaller $c_{v}$), as we expected, we observe a smaller KS but a larger MSE. 
\section{CONCLUSION}

This paper explores the connection between Singular Value Decomposition (SVD) and model fairness. We  construct a linear transformation for the weight matrix in neural networks and prove that the singular values of the SVD of the transformed weights are directly correlated with the disparities in the first and second moments of output distributions across sensitive social groups. Based on these findings, we propose the ESVDFair algorithm to enhance model fairness. This algorithm can be employed even when sensitive attributes are not available during the inference stage. Our approach shows a better or similar trade-off between accuracy and fairness when compared to state-of-the-art post-processing methods for fair regression. 

The proposed ESVDFair algorithm can be applied to any layer within a neural network. While our experiments focus on adjusting only the last and second-last layers, our findings motivate a potential future direction: identifying layers that contribute most significantly to unfairness and applying our algorithm to those specific layers. This could potentially improve the effectiveness of the proposed algorithm and further enhance the fairness-accuracy trade-off.

\section*{Acknowledgments}
This work is supported by the U.S. National Science Foundation under award
IIS-2301599 and CMMI-2301601, and by grants from the Ohio State University’s Translational Data
Analytics Institute and College of Engineering Strategic Research Initiative.

\bibliographystyle{unsrtnat}

\bibliography{paper}

\section*{Checklist}



 \begin{enumerate}

 \item For all models and algorithms presented, check if you include:
 \begin{enumerate}
   \item A clear description of the mathematical setting, assumptions, algorithm, and/or model. [Yes]
   \item An analysis of the properties and complexity (time, space, sample size) of any algorithm. [Yes]
   \item (Optional) Anonymized source code, with specification of all dependencies, including external libraries. [Yes]
 \end{enumerate}

 \item For any theoretical claim, check if you include:
 \begin{enumerate}
   \item Statements of the full set of assumptions of all theoretical results. [Yes]
   \item Complete proofs of all theoretical results. [Yes]
   \item Clear explanations of any assumptions. [Yes]     
 \end{enumerate}

 \item For all figures and tables that present empirical results, check if you include:
 \begin{enumerate}
   \item The code, data, and instructions needed to reproduce the main experimental results (either in the supplemental material or as a URL). [Yes]
   \item All the training details (e.g., data splits, hyperparameters, how they were chosen). [Yes]
         \item A clear definition of the specific measure or statistics and error bars (e.g., with respect to the random seed after running experiments multiple times). [Yes]
         \item A description of the computing infrastructure used. (e.g., type of GPUs, internal cluster, or cloud provider). [Yes]
 \end{enumerate}

 \item If you are using existing assets (e.g., code, data, models) or curating/releasing new assets, check if you include:
 \begin{enumerate}
   \item Citations of the creator If your work uses existing assets. [Yes]
   \item The license information of the assets, if applicable. [Not Applicable]
   \item New assets either in the supplemental material or as a URL, if applicable. [Not Applicable]
   \item Information about consent from data providers/curators. [Not Applicable]
   \item Discussion of sensible content if applicable, e.g., personally identifiable information or offensive content. [Not Applicable]
 \end{enumerate}

 \item If you used crowdsourcing or conducted research with human subjects, check if you include:
 \begin{enumerate}
   \item The full text of instructions given to participants and screenshots. [Not Applicable]
   \item Descriptions of potential participant risks, with links to Institutional Review Board (IRB) approvals if applicable. [Not Applicable]
   \item The estimated hourly wage paid to participants and the total amount spent on participant compensation. [Not Applicable]
 \end{enumerate}

 \end{enumerate}

\appendix

\onecolumn
\section{ADDITIONAL PROOFS}
\subsection{Proof for Lemma~\ref{eq:gaussian}}
\begin{proof}
    Because $\mathscr{X}_{a}^{[l]}$ follows a normal distribution, $\mathscr{Z}_{a}^{[l]} = \mathscr{X}_{a}^{[l]}(W^{[l]})^{T}$ also follows a normal distribution. When $\mathscr{Z}_{1}^{[l]}$ and $\mathscr{Z}_{2}^{[l]}$ have the same mean value and covariance matrix, they follow the same distribution. Because $\hat{\mathscr{Y}}|\mathscr{A} = 1$ and $\hat{\mathscr{Y}}|\mathscr{A} = 2$ are the functions of $\mathscr{Z}_{1}^{[l]}$ and $\mathscr{Z}_{2}^{[l]}$, they also follow the same distribution, which is to say that $\hat{\mathscr{Y}}$ is independent of $\mathscr{A}$.
\end{proof}

\subsection{Proof for Lemma~\ref{lemma1}}
\begin{proof}
    Since $(\bar{x}_{1} - \bar{x}_{2})^{T}(\bar{x}_{1} - \bar{x}_{2})$ is symmetric and positive semi-definite, $(\bar{x}_{1} - \bar{x}_{2})^{T}(\bar{x}_{1} - \bar{x}_{2}) + \epsilon_e I$ is positive definite matrix and has a Cholesky decomposition \cite{de1924equations}. Let $S_e$ be the Cholesky decomposition \cite{de1924equations} of $(\bar{x}_{1} - \bar{x}_{2})(\bar{x}_{1} - \bar{x}_{2})^{T} + \epsilon_e I$, $S_e$ satisfies Eq.~\ref{eq:lemma_1}. Since $S_e$ is a lower triangular matrix with positive diagonal entries, $S_{e}$ is invertible.
\end{proof}

\subsection{Proof for Theorem~\ref{the1}}
From the $S_{e}$ defined in the theorem, we have
\begin{align}
    \left\|\bar{x}_{1}W^{T} - \bar{x}_{2}W^{T}\right\|_{F}^{2} & = \mathrm{tr}\left[WS_{e}S_{e}^{-1}\left(\bar{x}_{1} - \bar{x}_{2}\right)^{T}\left(\bar{x}_{1} - \bar{x}_{2}\right)(S_{e}^{T})^{-1}S_{e}^{T}W^{T}\right] \\
    & = \mathrm{tr}\left[WS_{e}S_{e}^{-1}\left(S_{e}S_{e}^{T} - \epsilon_{e}I\right)(S_{e}^{T})^{-1}S_{e}^{T}W^{T}\right] \\
    & = \mathrm{tr}\left[WS_{e}(WS_{e})^{T}\right] - \epsilon_{e}\mathrm{tr}[WW^{T}].
\end{align}
Because
\begin{align}
    WS_{e} = \sum_{i=1}^{r_{e}}\sigma_{i(e)}u_{i(e)}v_{i(e)}^{T},
\end{align}
we have
\begin{align}
    \mathrm{tr}\left[WS_{e}(WS_{e})^{T}\right] = & = \mathrm{tr}\left[\left(\sum_{i=1}^{r_{e}}\sigma_{i(e)}u_{i(e)}v_{i(e)}^{T}\right)\left(\sum_{i=1}^{r_{e}}\sigma_{i(e)}u_{i(e)}v_{i(e)}^{T}\right)^{T}\right] \\
    & = \sum_{i=1}^{r_{e}}\sum_{j=1}^{r_{e}}\mathrm{tr}\left[\sigma_{i(e)}\sigma_{j(1)}u_{i(e)}v_{i(e)}^{T}v_{j(1)}u_{j(1)}^{T}\right] \\
    & = \sum_{i=1}^{r_{e}}\sigma_{i(e)}^{2}.
\end{align}
Therefore,
\begin{align}
    \left\|\bar{x}_{1}W^{T} - \bar{x}_{2}W^{T}\right\|_{F}^{2} = \sum_{i=1}^{r_{e}}\sigma_{i(e)}^{2} - \epsilon_{e} \mathrm{tr}[WW^{T}]
\end{align}
When
\begin{align}
    W'S_{e} = \sum_{i=1}^{r_{e}}\sigma_{i(e)}^{'}u_{i(e)}v_{i(e)}^{T},
\end{align}
\begin{align}
    \left\|\bar{x}_{1}W^{'T} - \bar{x}_{2}W^{'T}\right\|_{F}^{2} = \sum_{i=1}^{r_{e}}\sigma_{i(e)}^{'2} - \epsilon_{e} \mathrm{tr}[W'W^{'T}]
\end{align}

\subsection{Proof for Corollary~\ref{cor1}}
The element in the $i$-th row and $j$-th column of $WW^{T}$ is
\begin{align}
    (WW^{T})_{ij} =\sum_{k=1}^{n}w_{ik}w_{jk},
\end{align}
where $w_{ij}$ is the element in the $i$-th row and $j$-th column of $W$. So 
\begin{align}
    \mathrm{tr}[WW^{T}] = \sum_{i=1}^{m}\sum_{k=1}^{n}w_{ik}^{2} \geq 0.
\end{align}
So we have
\begin{align}
    \left\|\bar{x}_{1}W^{'T} - \bar{x}_{2}W^{'T}\right\|_{F}^{2} \leq \sum_{i=1}^{r_{e}}\sigma_{i(e)}^{'2}.
\end{align}

\subsection{A Lemma for Proof Theorem~\ref{the2}}
\begin{lemma}\label{lemma_abs_Lambda}
    For any matrix $W$, we have,  $$\left\|WMW^{T}\right\|_{F}^{2} = d_{v}^{2}(X_{1}, X_{2}; W)\leq \left\|W|M|W^{T}\right\|_{F}^{2}.$$
\end{lemma}

\subsection{Proof for Lemma~\ref{lemma_abs_Lambda}}
\begin{proof}
Because $M = Q\Lambda Q^{T}$, $WMW^{T} = WQ\Lambda Q^{T}W^{T}$. We denote $WQ = B$, then we have
\begin{align}
    WMW = B\Lambda B^{T}, ~~~~ W|M|W^{T} = B|\Lambda|B^{T}.
\end{align}
The element in $i$-th row and $j-$th column of $WMW$ is
\begin{align}
    (WMW^{T})_{ij} = \sum_{k=1}^{n}\lambda_{k}b_{ik}b_{jk}.
\end{align}
Therefore, we can get the squared Frobenius norm
\begin{align}
    \left\|WMW^{T}\right\|_{F}^{2} & = \sum_{i=1}^{m}\sum_{j=1}^{m} \left(\sum_{k=1}^{n}\lambda_{k}b_{ik}b_{jk}\right)^{2} \nonumber \\
    & = \sum_{i=1}^{m}\sum_{j=1}^{m}\sum_{k=1}^{n}\lambda_{k}^{2}b_{ik}^{2}b_{jk}^{2} + 2\sum_{i=1}^{m}\sum_{j=1}^{m}\sum_{k=1, l=1, k\neq l}\lambda_{k}\lambda_{l}b_{ik}b_{jk}b_{il}b_{jl}. 
\end{align}
Similarily, we have
\begin{align}
    \left\|W|M|W^{T}\right\|_{F}^{2} = \sum_{i=1}^{m}\sum_{j=1}^{m}\sum_{k=1}^{n}|\lambda_{k}|^{2}b_{ik}^{2}b_{jk}^{2} + 2\sum_{i=1}^{m}\sum_{j=1}^{m}\sum_{k=1, l=1, k\neq l}|\lambda_{k}||\lambda_{l}|b_{ik}b_{jk}b_{il}b_{jl}.     
\end{align}
The difference between them is
\begin{align}
    \left\|W|M|W^{T}\right\|_{F}^{2} - \left\|WMW^{T}\right\|_{F}^{2} = &2\sum_{i=1}^{m}\sum_{j=1}^{m}\sum_{k=1, l=1, k\neq l}|\lambda_{k}||\lambda_{l}|b_{ik}b_{jk}b_{il}b_{jl} - \nonumber \\ 
    & 2\sum_{i=1}^{m}\sum_{j=1}^{m}\sum_{k=1, l=1, k\neq l}\lambda_{k}\lambda_{l}b_{ik}b_{jk}b_{il}b_{jl}.
\end{align}
For every $k$ and $l$,
\begin{align}
    &\sum_{i=1}^{m}\sum_{j=1}^{m}|\lambda_{k}||\lambda_{l}|b_{ik}b_{jk}b_{il}b_{jl} - \sum_{i=1}^{m}\sum_{j=1}^{m}\lambda_{k}\lambda_{l}b_{ik}b_{jk}b_{il}b_{jl} \nonumber \\
    = & |\lambda_{k}||\lambda_{l}|\left(\sum_{i=1}^{m}b_{ik}b_{il}\right)\left(\sum_{j=1}^{m}b_{jk}b_{jl}\right) - \lambda_{k}\lambda_{l}\left(\sum_{i=1}^{m}b_{ik}b_{il}\right)\left(\sum_{j=1}^{m}b_{jk}b_{jl}\right) \nonumber \\
    = & \left(\sum_{i=1}^{m}b_{ik}b_{il}\right)^{2}\left(|\lambda_{k}||\lambda_{l}| - \lambda_{k}\lambda_{l}\right) \geq 0.
\end{align}
Therefore, $\left\|W|M|W^{T}\right\|_{F}^{2} - \left\|WMW^{T}\right\|_{F}^{2} \geq 0$. We have the Lemma~\ref{lemma_abs_Lambda} proved.
\end{proof}

\subsection{Proof for Theorem~\ref{the2}}
\begin{proof}
    From the property of $S_{v}$, we have
    \begin{align}
        \left\|W|M|W^{T}\right\|_{F}^{2} = \left\|WS_{v}S_{v}^{T}W^{T}\right\|_{F}^{2}.
    \end{align}
    Because
    \begin{align}
        WS_{v} = \sum_{i=1}^{r_{v}}\sigma_{i(v)}u_{i(v)}v_{i(v)}^{T}.
    \end{align}
    we have
    \begin{align}
        \left\|WS_{v}S_{v}^{T}W^{T}\right\|_{F}^{2} & = \mathrm{tr}\left[WS_{v}S_{v}^{T}W^{T}WS_{v}S_{v}^{T}W^{T}\right] \nonumber \\
        & = \mathrm{tr}\left[\left(\sum_{i=1}^{r_{v}}\sigma_{i(v)}u_{i(v)}v_{i(v)}^{T}\right)\left(\sum_{i=1}^{r_{v}}\sigma_{i(v)}v_{i(v)}u_{i(v)}^{T}\right)\left(\sum_{i=1}^{r_{v}}\sigma_{i(v)}u_{i(v)}v_{i(v)}^{T}\right)\left(\sum_{i=1}^{r_{v}}\sigma_{i(v)}v_{i(v)}u_{i(v)}^{T}\right)\right] \nonumber \\
        & = \sum_{i=1}^{r_{v}}\sum_{j=1}^{r_{v}}\sum_{k=1}^{r_{v}}\sum_{l=1}^{r_{v}} \sigma_{i(v)}\sigma_{j(v)}\sigma_{k(v)}\sigma_{l(v)} \mathrm{tr}\left[u_{i(v)}v_{i(v)}^{T}v_{j(v)}u_{j(v)}^{T}u_{k(v)}v_{k(v)}^{T}v_{l(v)}u_{l(v)}^{T}\right].
    \end{align}
    Because $v_{i(v)}^{T}v_{j(v)} = \delta_{ij}$, $v_{k(v)}^{T}v_{l(v)} = \delta_{kl}$,
    \begin{align}
        \left\|WS_{v}S_{v}^{T}W^{T}\right\|_{F}^{2} = \sum_{i=1}^{r_{v}}\sum_{k=1}^{r_{v}}\sigma_{i(v)}^{2}\sigma_{k(v)}^{2}\mathrm{tr}[u_{i(v)}u_{i(v)}^{T}u_{k(2)}u_{k(v)}^{T}].
    \end{align}
    Because $u_{i(v)}^{T}u_{k(v)} = \delta_{ij}$, $\mathrm{tr}[u_{i(v)}u_{i(v)}^{T}] = 1$,
    \begin{align}
        \left\|WS_{v}S_{v}^{T}W^{T}\right\|_{F}^{2} = \sum_{i=1}^{r_{v}}\sigma_{i(v)}^{4}.
    \end{align}

    For $W'$, when $S_{v}$ is invertible, we have
    \begin{align}
        W'S_{v} = \sum_{i=1}^{r_{v}}\sigma_{i(v)}^{'}u_{i(v)}v_{i(v)}^{T}.
    \end{align}
    Then
    \begin{align}
        \left\|W'S_{v}S_{v}^{T}W^{'T}\right\|_{F}^{2} = \sum_{i=1}^{r_{v}}\sigma_{i(v)}^{'4}.
    \end{align}
    When $S_{v}$ is not invertible, $S_{v}^{-1}$ is the pseudo-inverse of $S_{v}$,
    \begin{align}
        S_{v}^{-1}S_{v} = P,
    \end{align}
    where $P$ is a projection matrix to the row space of $S_{v}$. So,
    \begin{align}
        W'S_{v} = \left(\sum_{i=1}^{r_{v}}\sigma_{i(v)}^{'}u_{i(v)}v_{i(v)}^{T}\right)P.
    \end{align}
    Since 
    \begin{align}
        WS_{v} = \sum_{i=1}^{r_{v}}\sigma_{i(v)}u_{i(v)}v_{i(v)}^{T},
    \end{align}
    $\sum_{i=1}^{r_{v}}\sigma_{i(v)}u_{i(v)}v_{i(v)}^{T}$ lies entirely in the row space of $S_{v}$. When $\sigma_{i}^{'}$ are all non-zero values, $\sum_{i=1}^{r_{v}}\sigma_{i(v)}^{'}u_{i(v)}v_{i(v)}^{T}$ has the same row space of $\sum_{i=1}^{r_{v}}\sigma_{i(v)}u_{i(v)}v_{i(v)}^{T}$. When some of the $\sigma_{i}^{'}$ are 0, $\sum_{i=1}^{r_{v}}\sigma_{i(v)}^{'}u_{i(v)}v_{i(v)}^{T}$ lies in a subspace of $\sum_{i=1}^{r_{v}}\sigma_{i(v)}u_{i(v)}v_{i(v)}^{T}$. So, $\sum_{i=1}^{r_{v}}\sigma_{i(v)}^{'}u_{i(v)}v_{i(v)}^{T}$ also entirely lies in the row space of $S_{v}$, which means
    \begin{align}
        \left(\sum_{i=1}^{r_{v}}\sigma_{i(v)}^{'}u_{i(v)}v_{i(v)}^{T}\right)P = \left(\sum_{i=1}^{r_{v}}\sigma_{i(v)}^{'}u_{i(v)}v_{i(v)}^{T}\right).
    \end{align}
    So, we have
    \begin{align}
        W'S_{v} = \left(\sum_{i=1}^{r_{v}}\sigma_{i(v)}^{'}u_{i(v)}v_{i(v)}^{T}\right),
    \end{align}
    therefore
    \begin{align}
        \left\|W'S_{v}S_{v}^{T}W^{'T}\right\|_{F}^{2} = \sum_{i=1}^{r_{v}}\sigma_{i(v)}^{'4}.
    \end{align}
\end{proof}

\subsection{Proof for Theorem~\ref{the:opt1}}
The objective function can be written as
{\scriptsize
\begin{align}
    \left\|XW_{e}^{'T} - XW^{T}\right\|_{F}^{2} & = \left\|X(S_{e}^{-1})^{T}\left(\sum_{i=1}^{r_{e}}\sigma_{i(e)}^{'}v_{i(e)}u_{i(e)}^{T} - \sum_{i=1}^{r_{e}}\sigma_{i(e)}v_{i(e)}u_{i(e)}^{T}\right)^{T}\right\|_{F}^{2} \\ 
    & = \mathrm{tr}\left[X(S^{-1})^{T}\left(\sum_{i=1}^{r_{e}}\sigma_{i(e)}^{'}v_{i(e)}u_{i(e)}^{T} - \sum_{i=1}^{r_{e}}\sigma_{i(e)}v_{i(e)}u_{i(e)}^{T}\right)^{T}\left(\sum_{i=1}^{r_{e}}\sigma_{i(e)}^{'}v_{i(e)}u_{i(e)}^{T} - \sum_{i=1}^{r_{e}}\sigma_{i(e)}v_{i(e)}u_{i(e)}^{T}\right)(S_{e}^{-1})X^{T}\right] \\
    & = \mathrm{tr}\left[\left(\sum_{i=1}^{r_{e}}\sigma_{i(e)}^{'}u_{i(e)}v_{i(e)}^{T} - \sum_{i=1}^{r_{e}}\sigma_{i(e)}u_{i(e)}v_{i(e)}^{T}\right)(S_{e}^{-1})X^{T}X(S_{e}^{-1})^{T}\left(\sum_{i=1}^{r_{e}}\sigma_{i(e)}^{'}v_{i(e)}u_{i(e)}^{T} - \sum_{i=1}^{r_{e}}\sigma_{i(e)}v_{i(e)}u_{i(e)}^{T}\right)\right].
\end{align}
}
Consider a single item,
\begin{align}
    \mathrm{tr}\left[\sigma_{i(e)}^{'}u_{i(e)}v_{i(e)}^{T}(S_{e}^{-1})X^{T}X(S_{e}^{-1})^{T}\sigma_{j(1)}^{'}v_{j(1)}u_{j(1)}^{T}\right] = \delta_{ij}\sigma_{i(e)}^{'}\sigma_{j(1)}^{'}v_{i(e)}^{T}(S_{e}^{-1})X^{T}X(S_{e}^{-1})^{T}v_{(1)j}.
\end{align}
Define
\begin{align}
    k_{i(e)} = v_{i(e)}^{T}(S_{e}^{-1})X^{T}X(S_{e}^{-1})^{T}v_{i(e)},
\end{align}
the objective function is
\begin{align}
    \left\|XW_{e}^{'T} - XW^{T}\right\|_{F}^{2} = \sum_{i}^{r_{e}}k_{i(e)}\sigma_{i(e)}^{'2} - 2\sum_{i=1}^{r_{e}}\sigma_{i(e)}^{'}\sigma_{i(e)}k_{i(e)} + \sum_{i=1}^{r_{e}}k_{i(e)}\sigma_{i(e)}^{2}.
\end{align}
Then we can write the Lagrange function
\begin{align}
    \mathcal{L}(\sigma_{i(e)}^{'}, \gamma_{e}) = \sum_{i(e)}^{r_{e}}k_{i(e)}\sigma_{i(e)}^{'2} - 2\sum_{i=1}^{r_{e}}\sigma_{i(e)}^{'}\sigma_{i(e)}k_{i(e)} + \sum_{i=1}^{r_{e}}k_{i(e)}\sigma_{i(e)}^{2} + \lambda\left(\sum_{i=1}^{r_{e}}\sigma_{i(e)}^{'2} - c_{e}\right).
\end{align}
Because
\begin{align}
    \frac{\partial \mathcal{L}(\sigma_{i(e)}^{'}, \gamma_{e})}{\partial \sigma_{i(e)}^{'}} = 2k_{i(e)}\sigma_{i(e)}^{'} - 2\sigma_{i(e)}k_{i(e)} + 2\gamma_{e} \sigma_{i(e)}^{'} = 0,
\end{align}
\begin{align}
    \frac{\partial \mathcal{L}(\sigma_{i(v)}^{'}, \gamma_{v})}{\partial \gamma_{e}} = \sum_{i=1}^{r_{e}}\sigma_{i(e)}^{'2} - c_{e} = 0,
\end{align}
we can get the solution
\begin{align}
    \sigma_{i(e)}^{'} = \frac{\sigma_{i(e)}k_{i(e)}}{k_{i(e)} + \gamma_{e}}.
\end{align}
$\gamma_{e}$ is the solution of the equation
\begin{align}
    \sum_{i=1}^{r_{e}}\sigma_{i(e)}^{'2} = c_{e}.
\end{align}

\subsection{Proof for Theorem~\ref{the:opt2}}
Similar to the proof of Theorem~\ref{the:opt1}, the objective function can be written as
\begin{align}
    \left\|XW_{v}^{'T} - XW^{T}\right\|_{F}^{2} = \sum_{i}^{r_{v}}k_{i(v)}\sigma_{i(v)}^{'2} - 2\sum_{i=1}^{r_{v}}\sigma_{i(v)}^{'}\sigma_{i(v)}k_{i(v)} + \sum_{i=1}^{r_{v}}k_{i(v)}\sigma_{i(v)}^{2},
\end{align}
where
\begin{align}
    k_{i(v)} = v_{i(v)}^{T}(S_{v}^{-1})X^{T}X(S_{v}^{-1})^{T}v_{i(v)}.
\end{align}
We define the Lagrange function as 
\begin{align}
    \mathcal{L}(\sigma_{i(v)}^{'}, \gamma_{v}) = \sum_{i=1}^{r_{v}}k_{i(v)}\sigma_{i(v)}^{'2} - 2\sum_{i=1}^{r_{v}}\sigma_{i(v)}^{'}\sigma_{i(v)}k_{i} + \sum_{i=1}^{r_{v}}k_{i}\sigma_{i(v)}^{2} + \gamma_{v}(\sum_{i=1}^{r_{v}}\sigma_{i(v)}^{'4} - c_{v}).
\end{align}
Then we have
\begin{align}
    \frac{\partial \mathcal{L}(\sigma_{i(v)}, \gamma_{v})}{\partial \sigma_{i(v)}^{'}} = 2k_{i(v)}\sigma_{i(v)}^{'} - 2k_{i(v)}\sigma_{i(v)} + 4\gamma_{v}\sigma_{i(v)}^{'3},
\end{align}
\begin{align}
    \frac{\partial \mathcal{L}(\sigma_{i(v)}, \gamma_{v})}{\partial \gamma_{v}} = \sum_{i=1}^{r_{v}}\sigma_{i(v)}^{'4} - c_{v}.
\end{align}
Solving the equation
\begin{align}
    2k_{i(v)}\sigma_{i(v)}^{'} - 2k_{i(v)}\sigma_{i(v)} + 4\gamma_{v}\sigma_{i(v)}^{'3} = 0,
\end{align}
we can get
{\small
\begin{align}
    \sigma_{i(v)}^{'} = -\frac{k_{i(v)}}{6^{\frac{1}{3}}\left(9\gamma_{v}^{2}k_{i(v)}\sigma_{i(v)} + \sqrt{3}\sqrt{2\gamma_{v}^{3}k_{i(v)}^{3} + 27\gamma_{v}^{4}k_{i(v)}\sigma_{i(v)}^{2}}\right)^{\frac{1}{3}}} + \frac{\left(9\gamma_{v}^{2}k_{i(v)}\sigma_{i(v)} + \sqrt{3}\sqrt{2\gamma_{v}^{3}k_{i(v)}^{3} + 27\gamma_{v}^{4}k_{i(v)}\sigma_{i(v)}^{2}}\right)^{\frac{1}{3}}}{6^{\frac{2}{3}}\gamma_{v}}.
\end{align}
}
$\gamma_{v}$ is the solution of the equation
\begin{align}
    \sum_{i=1}^{r_{v}}\sigma_{i(v)}^{'2} = c_{v}.
\end{align}

\section{ARCHITECTURE OF PRE-TRAINED PREDICTOR}
Table~\ref{tab:nn_architecture} displays the architecture of the neural network we used in the experiment. It consists of 5 linear layers. The first four layers are followed by a ReLU activation function. 

For this network, our method only involves doing svd for a matrix with the shape $256 \times 256$. In general, the most complexity for our method come through the matrix multiplication of $(X^{[l]})^{T}X^{[l]}$. Since $X \in \mathbb{R}^{N \times n^{[l]}}$, it is $\mathcal{O}(N^{2}n^{[l]2})$.

\begin{table}[htbp]
    \centering
    \caption{The Neural Network Architecture used as the unfair base predictor in the experiments.}\label{tab:nn_architecture}
    \begin{tabular}{ccc}
    \toprule
    Layer Type & Input Size & Output Size  \\
    \midrule
    Linear & -- & 256 \\
    ReLU activation & -- & -- \\
    Linear & 256 & 256 \\
    ReLU activation & -- & -- \\
    Linear & 256 & 256 \\
    ReLU activation & -- & -- \\
    Linear & 256 & 256 \\
    ReLU activation & -- & -- \\
    Linear & 256 & 1 \\
    \bottomrule
    \end{tabular}
\end{table}

\section{DENSITY OF THE OUTPUT DISTRIBUTION WITH DIFFERENT $c_{2}$}
\begin{figure}[htbp]
    \centering
    \begin{subfigure}[b]{0.19\textwidth}
        \includegraphics[width=\textwidth]{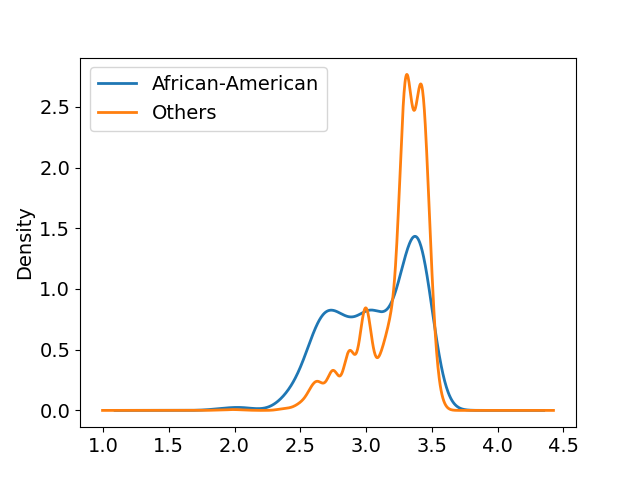}
        \caption{$\tilde{c}_{v} = 5$}\label{fig:law_distribution_c_2_5}
    \end{subfigure}
    \begin{subfigure}[b]{0.19\textwidth}
        \includegraphics[width=\textwidth]{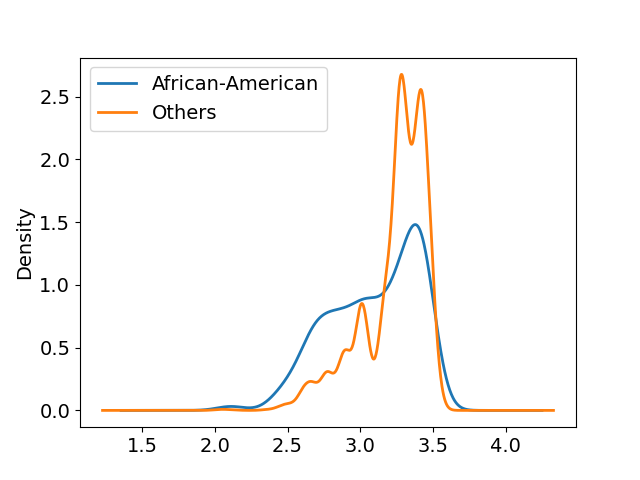}
        \caption{$\tilde{c}_{v} = 10$}\label{fig:law_distribution_c_2_10}
    \end{subfigure}
    \begin{subfigure}[b]{0.19\textwidth}
        \includegraphics[width=\textwidth]{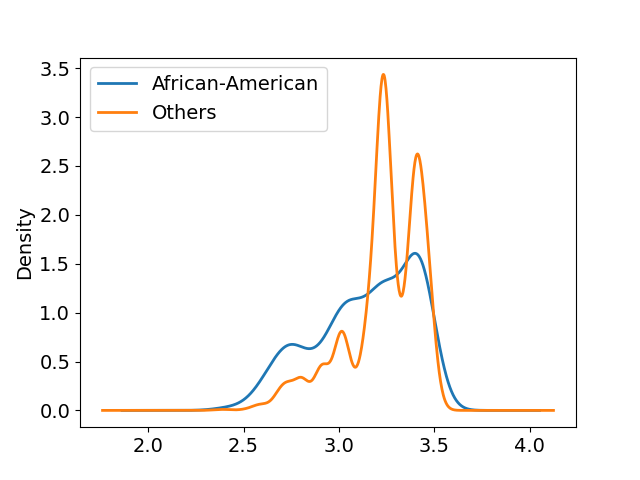}
        \caption{$\tilde{c}_{v} = 50$}\label{fig:law_distribution_c_2_50}
    \end{subfigure}
    \begin{subfigure}[b]{0.19\textwidth}
        \includegraphics[width=\textwidth]{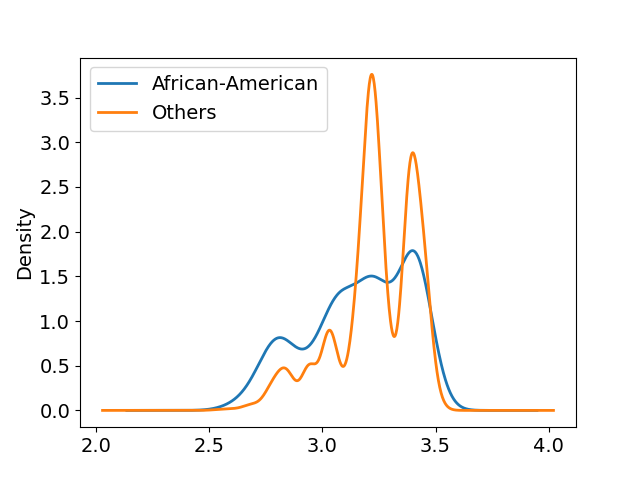}
        \caption{$\tilde{c}_{v} = 100$}\label{fig:law_distribution_c_2_100}
    \end{subfigure}
    \begin{subfigure}[b]{0.19\textwidth}
        \includegraphics[width=\textwidth]{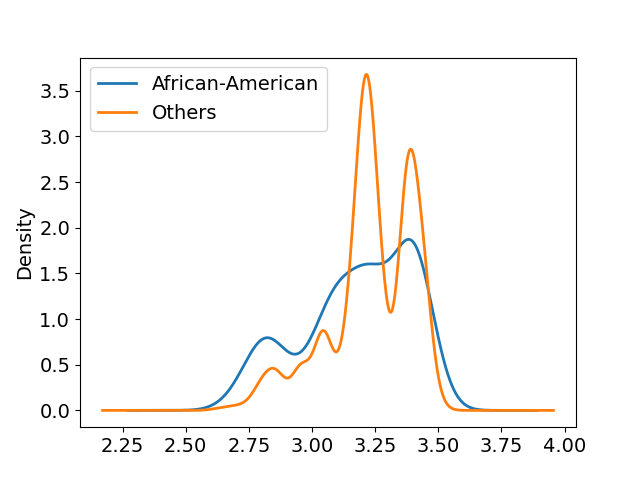}
        \caption{$\tilde{c}_{v} = 150$}\label{fig:law_distribution_c_2_150}
    \end{subfigure}
    \caption{Density of the output distribution across two sensitive groups on Law School Sucess dataset with different $\tilde{c}_{v}$.}\label{fig:law_distribution_c_2}
\end{figure}

\end{document}